\documentclass{article}

   \usepackage[preprint]{tackling_climate_workshop_style}

\usepackage[utf8]{inputenc} 
\usepackage[T1]{fontenc}    
\usepackage{hyperref}       
\usepackage{url}            
\usepackage{booktabs}       
\usepackage{amsfonts}       
\usepackage{nicefrac}       
\usepackage{microtype}      
\usepackage{amsmath}
\usepackage{algorithmic}
\usepackage{array}
\usepackage[tight,footnotesize]{subfigure}
\usepackage{url}
\usepackage{accents}
\usepackage{booktabs}
\usepackage[table,xcdraw]{xcolor}
\usepackage[pdftex]{graphicx}

\title{\textit{Efficient Training of Physics-Informed Neural Networks with Direct Grid
Refinement Algorithm}} 

\author{%
  Shikhar Nilabh, Fidel Grandia \\
  Amphos 21 Consulting S.L.,\\
  Barcelona, Spain\\
  \texttt{{shikhar.nilabh, fidel.grandia}@amphos21.com} \\
}

\begin{document}

\maketitle

\begin{abstract}
This research presents the development of an innovative algorithm tailored for the adaptive sampling of residual points within the framework of Physics-Informed Neural Networks (PINNs). By addressing the limitations inherent in existing adaptive sampling techniques, our proposed methodology introduces a direct mesh refinement approach that effectively ensures both computational efficiency and adaptive point placement. Verification studies were conducted to evaluate the performance of our algorithm, showcasing reasonable agreement between the model based on our novel approach and benchmark model results. Comparative analyses with established adaptive resampling techniques demonstrated the superior performance of our approach, particularly when implemented with higher refinement factor. Overall, our findings highlight the enhancement of simulation accuracy achievable through the application of our adaptive sampling algorithm for Physics-Informed Neural Networks.

\end{abstract}

\section{Introduction}

Physics-informed neural networks (PINNs) have gained prominence in recent years as a versatile tool for solving partial differential equations (PDEs) governed problems using deep neural networks (DNNs). Although PINNs have demonstrated success, addressing a broad range of increasingly complex PDE problems presents theoretical and practical challenges, necessitating further advancements to enhance prediction accuracy, computational efficiency, and training robustness [1].
Various techniques, such as loss function meta-learning [2], gradient-enhanced PINN [3], and adaptive sampling of residual points [4,5] have been employed to enhance the accuracy of PINNs. Our focus is on improving PINN accuracy through a novel algorithm for adaptive non-uniform sampling. Two common approaches for adaptive sampling methods (ASM) are identified [4]. The first approach (ASM 1) selects points from the original residual set based on a probability mass function (PMF) [6].
Although computationally efficient,  ASM 1 only selects the additional point only from the existing set of residual points. Therefore, it does not introduce any new points at different locations within the input space. Previous research has shown that the adaptive location of the resampled points further enhances the accuracy of PINNs [7]. 
An alternative approach to Adaptive Sampling, reffered as ASM2, considers addition of residual points at new locations within the input space [5]. In ASM2, a random sampling of residual points takes place over the input space, and those with relatively higher PDE residual values are selected. This method facilitates the adaptive positioning of residual points and is intutionally similar to adaptive mesh refinement technique used in numerical methods. However, ASM 2 is  computationally expensive as it requires calculation of PDE residual at all randomly chosen points during each resampling period [8]. It is worth noting that recent research studies have introduced various variants of these two adaptive sampling methods [9,10,11, and 12].
In this research, we introduce a novel adaptive sampling scheme for sampling points from new locations in the input space based on the PDE residual of original residual points (ASM 3). The scheme consists of three steps. In Step 1, an equi-spaced grid of residual points is defined as the reference set throughout the training process. In Step 2, new points are sampled from the reference residual points at each resampling period using their probability distribution function (as in ASM 1). In Step 3, a new set of points is added in the neighborhood of each sampled point from Step 2. For a refinement factor of 2, one point is added, while for higher refinement factors, multiple points are added. The mathematical definition of the neighborhood in Step 3 is provided in Section 2.
The proposed method exhibits computational efficiency by utilizing the PDE residual on the reference residual points in Step 2, eliminating the need for additional calculations. This efficiency enables a higher frequency of resampling, thereby increasing accuracy. Furthermore, the method achieves adaptive point placement by assigning new points in the vicinity of the sampled points from Step 2, akin to the mesh refinement technique used in numerical studies. In contrast to ASM2, which indirectly refines the grid and is independent of the original set of residual points, this algorithm directly refines the grid formed by the reference set of residual points in Step 1. Additionally, like adaptive mesh refinement, it offers flexibility in assigning higher refinement for improved PINN accuracy. Notably,
in order to enhance computational efficiency of ASM3 algorithm, previously sampled points in Step 3 are not retained or utilized in the subsequent resampling events.

\section{ Direct Grid Refinement Method}
For the development of ASM 3, we considered a transient PDE case with $t\in[0,T]$ and $x\in \Omega $ (where $\Omega \in R^D$). In step 1, a set of uniformly spaced residual points $\left \{ {t_f^i,x_f^i } \right \}_{i=1}^{N_f}$ are defined with a corresponding grid size of $h_t$  and $h_x$. At each resampling period, m points $\left \{  { t_s^i,x_s^i }\right \}_{i=1}^m $   are sampled from the set of reference residual points based on a probability mass function (elaborated in section (\ref{sec2.1})). For a refinement factor of 2, a new point $\left \{ {t_r^i,x_r^i } \right \}_{i=1}^m$ is selected in the neighborhood of each sampled point  $\left ( t_s^i,x_s^i \right )$ in step 2. The location of the refined point  $(t_r^i,x_r^i)$ on the input space is represented by Equation \ref{eq1} and \eqref{eq2}:

\begin{equation}
t_r^i= t_s^i+ \lambda_t h_t
\label{eq1}
\end{equation}

\begin{equation}
x_r^i= x_s^i+ \lambda_x h_x
\label{eq2}
\end{equation}
where $\left \{ \lambda_t,\lambda_x  \right \}$  are the refinement coefficients which range between $-1$ to $1$. These coefficients can be assigned as constants or dynamically selected by randomly picking values between $-1$ and $1$ for each sampled point in Step 2. Equation \eqref{eq1} and \eqref{eq2} illustrate the addition of a single point in the neighborhood of the sampled point from Step 2, resulting in a refinement factor of $2$. It is also possible to achieve higher refinement orders by adding multiple refined points using different values of the refinement coefficients.
\begin{figure}[!h]
\vskip 0.2in
\begin{center}
\centerline{\includegraphics[width=\columnwidth]{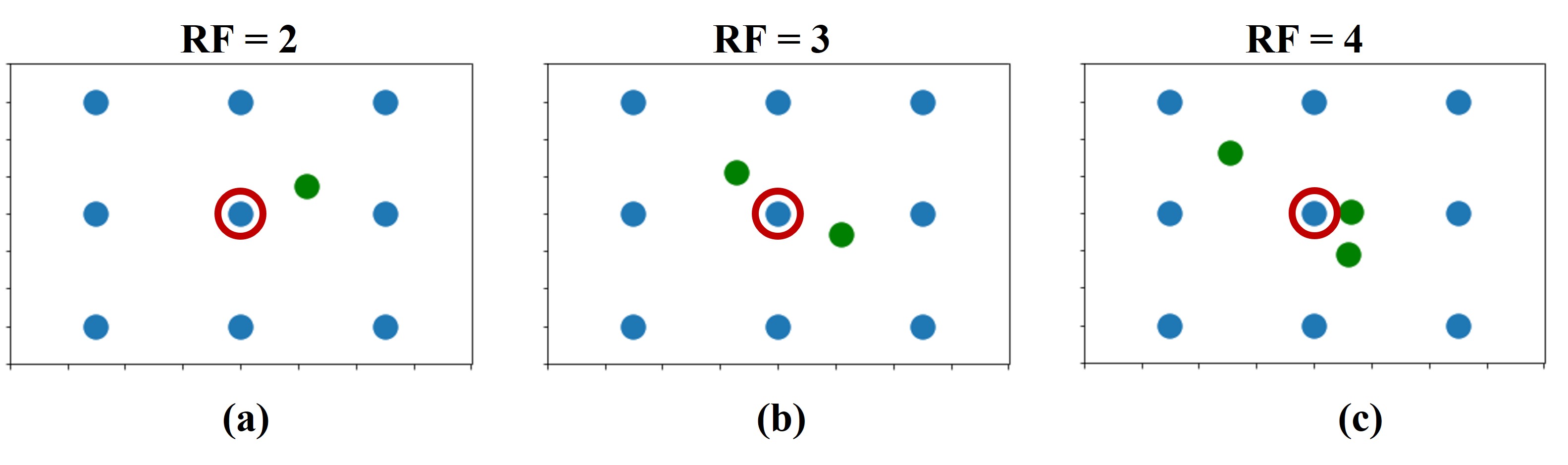}}
\caption{Implementation of ASM 3 with refinement factor of 2, 3 and 4. The blue dots represent reference residual points defined in Step 1, the blue dot encircled by red circle represent the sampled point from a PMF of reference residual points. The green points represent the addition of the new residual points by the direct grid refinement algorithm (Step 3).}
\label{fig1}
\end{center}
\vskip -0.2in
\end{figure}
Fig. \eqref{fig1} depicts the implementation of ASM3 with refinement factors of 2, 3, and 4, considering randomized refinement coefficients. The blue dots represent the reference residual points defined in Step 1, while the blue dot encircled by a red circle represents the sampled point from the PMF of reference residual points in Step 2. The green dots represent the adaptively sampled points in Step 3.

\subsection{Probability Mass function}
\label{sec2.1}
The probability mass function used in Step 2 of ASM and is represented by Equation \eqref{eq3}:
\begin{equation}
\overline{p}\left ( x \right )=\frac{p(x)}{\sum_{x\in S_0} p(x)}
\label{eq3}
\end{equation}

where $x(x\in S_0)$ represents a point from the reference set of equi-distant residual points $(S_0)$. $p(x)$ is the probability density function which is a non-linear function of PDE residual. [4] defined a general expression for PDF using k and c as hyperparameters (Equation \eqref{eq4}). 

\begin{equation}
p(x) 	\propto \frac{\epsilon^k(x)}{\mathbb{E}[\epsilon^k(x)]} +c
\label{eq4}
\end{equation}
where  $\epsilon(x)$ is the PDE residual at residual point $ x$.  In this research work, the value of k and c is taken as 2 and 0 respectively, a combination which has already been tested in a previous research work [3].

\subsection{Solving Advection Dispersion Equation }
A non-linear transient PDE equation governing the advection-dispersion process in a 1 dimensional porous medium is solved with ASM 3 (Equation \eqref{eq5},\eqref{eq6},\eqref{eq7}):
\begin{equation}
\frac{\partial \left ( \epsilon c \right )}{\partial t}= - \frac{\partial \left ( \nu  c \right )}{\partial x}+ \frac{\partial \left ( D_e +\alpha_L \nu \right )\frac{\partial c}{\partial x}}{\partial x} 
\label{eq5}
\end{equation}
where $ x \in\left [ 0,1 \right ], t \in [0,6000]$
\begin{equation}
c\left ( t,0 \right ) = \left ( \frac{1}{1+e^{-0.02\left ( t-500 \right )}} \right ) \times \left ( \frac{1}{1+e^{0.02\left ( t+500 \right )}} \right ) 
\label{eq6}
\end{equation}

\begin{equation}
\left ( D_e+\alpha_L  \nu\right )\frac{\partial_c\left ( t,1 \right )}{\partial x} = 0
\label{eq7}
\end{equation}
where $\epsilon$ is the porosity, $D_e$ is molecular dispersivity $(m^2\cdot s^{-1}), \alpha_L$ dispersivity (m).  Table \ref*{tab1} enlists the properties of the porous medium  for the model development.

\begin{table}[!ht]
\caption{List of porous medium properties and their values used for model development.}
\label{tab1}
\vskip 0.15in
\begin{center}
\begin{small}
\begin{sc}
\begin{tabular}{lc}
\toprule
Properties & Value \\
\midrule
 porosity                        & 0.3            \\
dispersivity $(m) $               & 0.01           \\
Groundwater velocity $(m\cdot s^{-1})$    & 0.0003         \\
Dispersion coefficient $(m^2\cdot s^{-1})$ & $1 \times 10^{-9 }   $      \\
\bottomrule
\end{tabular}
\end{sc}
\end{small}
\end{center}
\vskip -0.1in
\end{table}

The network output is $c(t,x)$ which represent the concentration of a non-reactive chemical species in the porous medium. The entry of this species in the medium is mathematically smoothed using a sigmoidal step function.
The other boundary condition, $(c(t, 1) = 0)$, represents an open boundary of the exit of chemical species at the other end of the porous medium. The porous medium is considered to be free of the chemical species and thus $c(0, x )$ is considered to be 0. 
The neural network consists of 3 hidden layers with 50 neurons each, utilizing sigmoid activation function for non-linearity. A total of 202 residual points enforce the boundary conditions, while 441 residual points (equally spaced) enforce equation \eqref{eq5}. The study utilizes a weighted-loss function, as formulated in a previous research work [13].

For the comparative study, four different simulations are run for each of the adaptive sampling methods, namely ASM 1, ASM 2, ASM 3. The model runs for 15000 steps of gradient descent (using Adam optimizer) with resampling period of 1000 epoch. For each method, 150 new points are added at each resampling event. Two variants of ASM 3 are implemented, one with RF =2 (ASM 3a) and RF = 4 (ASM 3b). In addition, a simulation with fixed number of residual points (without a sampling strategy) is also simulated for the base case study.

\section{Results}
\subsection{Model Verification}
\begin{figure}[!ht]
\vskip 0.2in
\begin{center}
\centerline{\includegraphics[width=\columnwidth]{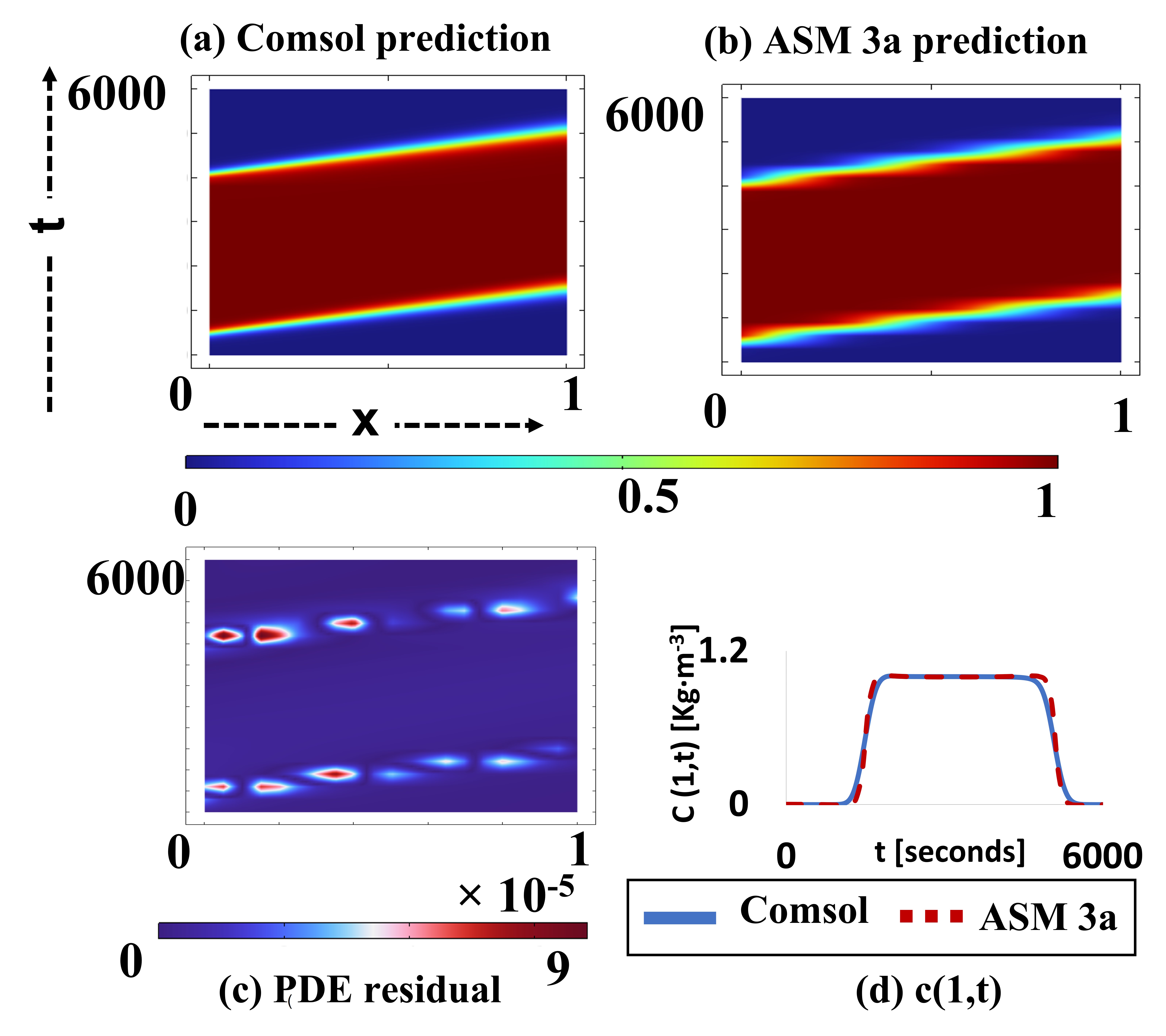}}
\caption{(a) Benchmark model result from Comsol used for the verification study, (b) Result from PINN using ASM 3a sampling method (c) Error analysis of the reference set of residual points highlights relatively higher error at a region of higher concentration gradient, (d) Comparative analysis of the PINN model result and the Comsol model result for the chemical species concentration  $c(t,1)$ shows excellent match with a $R^2$ value of 99.84\% .}
\label{fig2}
\end{center}
\vskip -0.2in
\end{figure}

The accuracy of the novel direct grid refinement algorithm is evaluated using a benchmark model implemented in Comsol. Fig. \ref{fig2} (a) displays the concentration profile $c(t,x)$ predicted by the Comsol benchmark model. The direct grid refinement algorithm, after 15000 epochs, closely approximates the Comsol model with minor discrepancies, as shown in Fig. \ref{fig2} (b). 

\begin{figure}[!ht]
\vskip 0.2in
\begin{center}
\centerline{\includegraphics[width=\columnwidth]{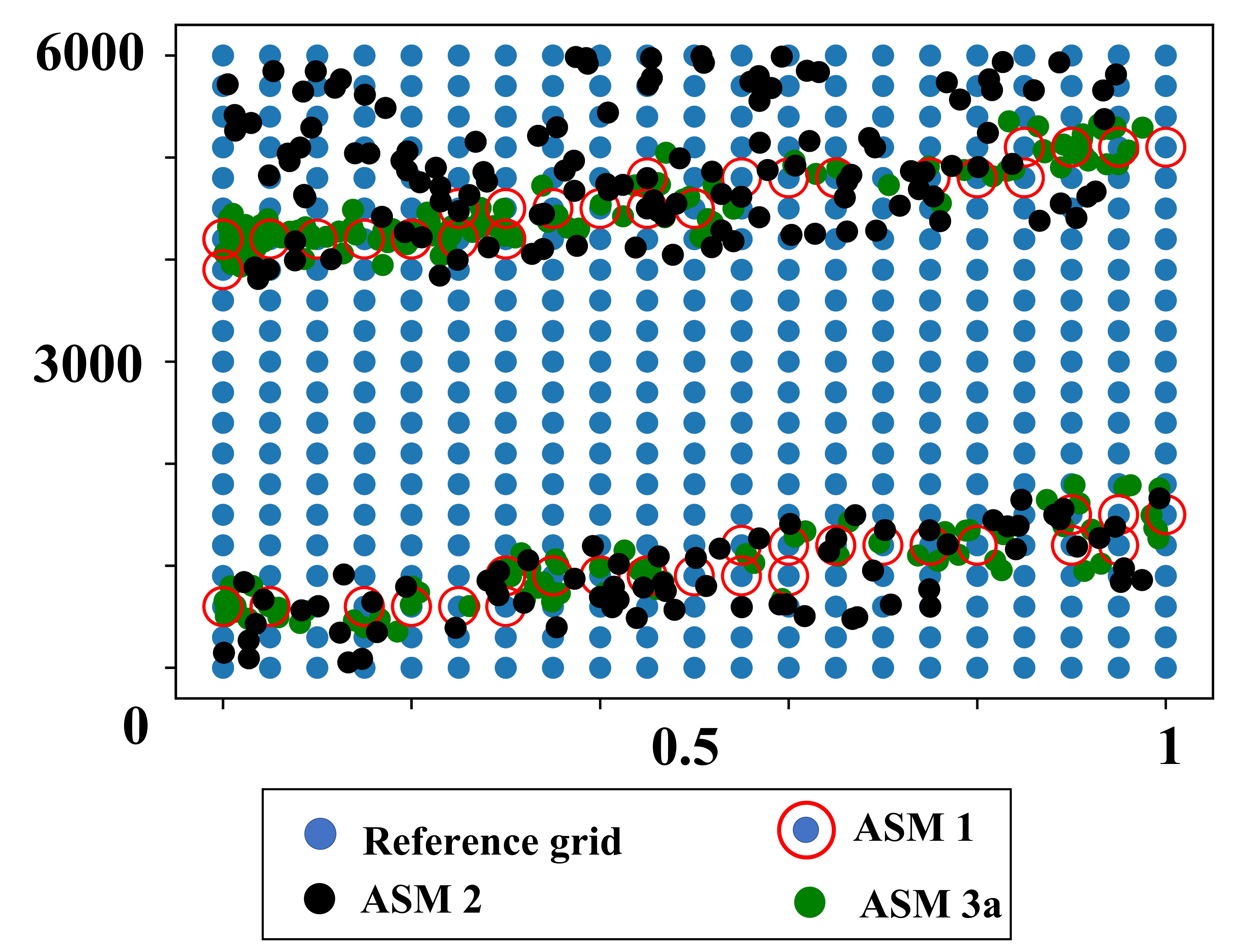}}
\caption{After 15000 epochs, the reference residual points (blue color) along with the sampled points from ASM1, 2 and 3 (red encircled blue dots, black dots and green dots respectively).}
\label{fig3}
\end{center}
\vskip -0.2in
\end{figure}

The discrepancies primarily occur in the region with sharp concentration gradients. These discrepancies are further illustrated in the error analysis, where the plot of PDE residual (Fig. \ref{fig2} (c)) reveals an error ranging up to $9 \times 10^{-5}$ at the front of the concentration gradient. For the verification of ASM 3a against the comsol result, the  concentration profile at $x=1$ is plotted for all time steps (Fig. \ref{fig2} (d)). The high $R^2$ value of $99.84\%$ demonstrates the efficiency of the PINN utilizing the novel sampling algorithm.

\begin{figure}[!ht]
\vskip 0.2in
\begin{center}
\centerline{\includegraphics[width=\columnwidth]{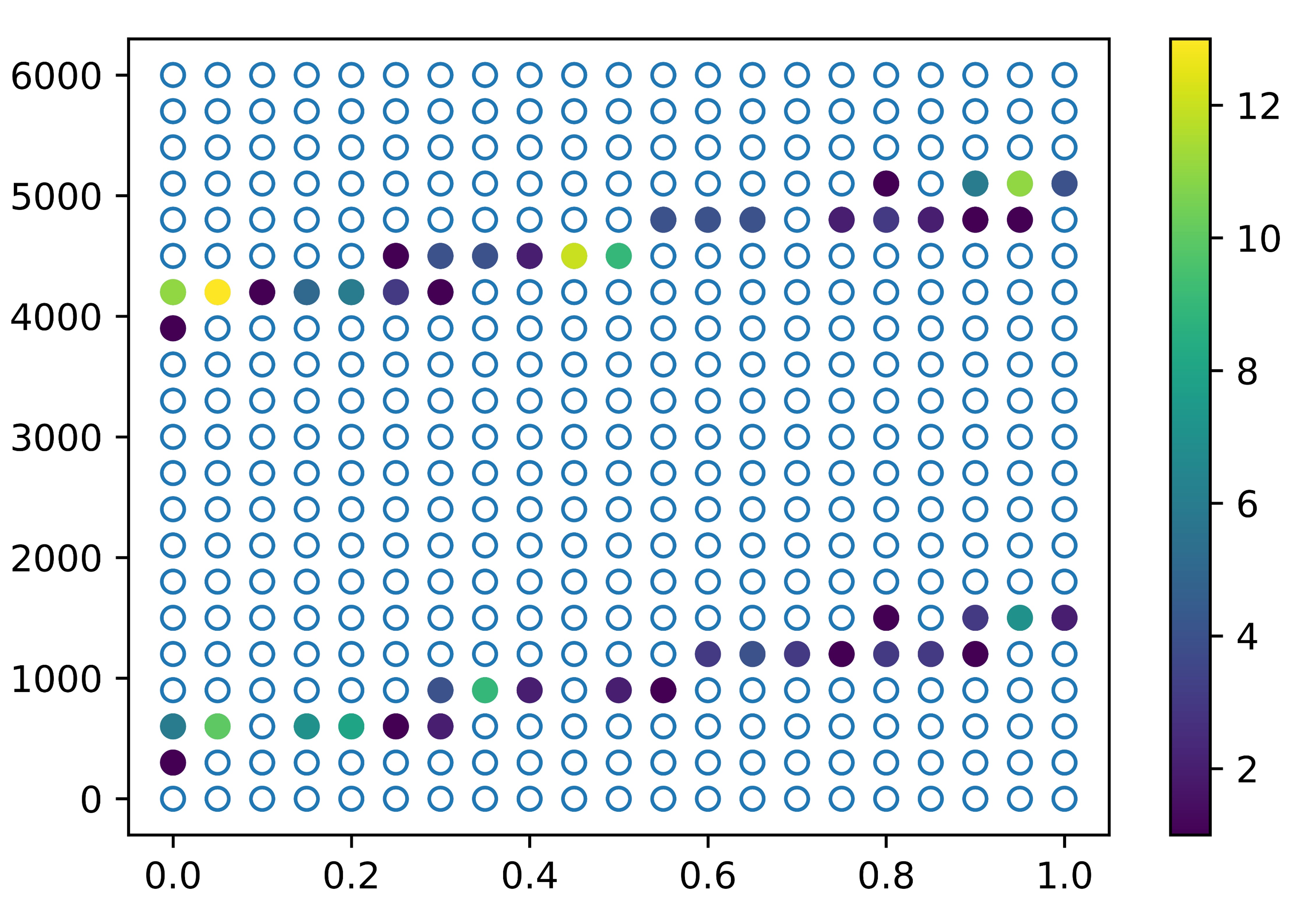}}
\caption{Sampling of residual points in Step 1 and Step 2 of ASM 3a. The empty blue circles represent the points from the set of reference grid defined in Step 1 of ASM 3a. The color filled circle represents the sampled points in the Step 2 of ASM 3a. The color map represents the repetition frequency of each point sampled in Step 2.}
\label{fig4}
\end{center}
\vskip -0.2in
\end{figure}

Fig. \ref{fig3}  depicts the adaptively sampled point using the grid refinement algorithm (ASM 3a), represented by a green dots. The alignment of the green dots with the regions of high PDE residual Fig. \ref{fig2} (c) is evident. Additionally, for comparison, the adaptively sampled points from ASM 1 and ASM 2 are also plotted. Fig. \ref*{fig3}  illustrates that the majority of adaptively sampled points lie in the region with relatively higher residual error. ASM 1 samples from set of original residual points, as indicated by the red encircled residual points, while ASM 2 selects randomized residual points with higher error.  

The adaptively sampled points in ASM 3a exhibit dense clusters compared to the residual points sampled with the ASM 1 method, as shown in Fig. \ref{fig3}. This observation can be attributed to the repetitive sampling of residual points in the second step of ASM 3a, similar to the case of ASM 1. To investigate the repetitive sampling, a color map representing the sampled points from the probability density function (PDF) of the reference grid is plotted Fig. \ref{fig4}. Each resampling event adds 150 new points, and the repetition of certain points occurs during the sampling process of Step 2 in ASM 3a (indicated by filled circles). The color map represents the frequency of point repetition, ranging from 1 to 13. The higher repetition of sampled points in the steps contributes to the clustering of adaptively sampled points in Step 3, as observed in Fig. \eqref{eq3}.

\subsection{Comparative Study}
The efficiency of different resampling methods is evaluated by assessing their relative L2 error  in comparison to the Comsol-benchmark model. 
As illustrated in the Table \ref*{tab2}, the results indicate that the adaptive sampling techniques exhibit higher accuracy levels than the base case (PINN without adaptive resampling scheme). The model based on the novel algorithm (ASM 3a) demonstrates relatively superior accuracy when compared to ASM 1, as evidenced by the associated relative L2 error. However, in contrast to ASM 2, the novel algorithm-based model (ASM 3a) exhibits a relatively higher relative L2 error. With refinement factor of 4, the accuracy corresponding to the novel algorithm (ASM 3b) improves further and outperforms the other adaptive sampling techniques.

\begin{table}[!ht]
\caption{Efficiency of PINN using different sampling techniques using relative L2 error.}
\label{tab2}
\vskip 0.15in
\begin{center}
\begin{small}
\begin{sc}
\begin{tabular}{lc}
\toprule
ASM & Relative L2 error \\
\midrule
\begin{tabular}[c]{@{}l@{}}Base   case \end{tabular} & $4.01 \times 10^{-4}$                          \\
ASM 1                          & 2.80$\times 10^{-4}$                         \\
ASM 2                          & 2.38$\times 10^{-4}$                           \\
ASM $3a$                         & 2.51$\times10^{-4} $                            \\
ASM $3b$                         & 2.09$\times 10^{-4} $  \\

\bottomrule
\end{tabular}
\end{sc}
\end{small}
\end{center}
\vskip -0.1in
\end{table}

\section{Conclusion}
 
In this study, we developed a novel algorithm for adaptive sampling of residual points. Verification studies using a Physics-Informed Neural Network (PINN) with our sampling method (ASM 3a) demonstrated reasonable agreement with benchmark model results. Comparative analysis with other techniques revealed that our approach delivered satisfactory results, especially with higher refinement factors. Overall, our adaptive sampling algorithm enhances simulation accuracy and efficiency, offering promising prospects for future research and applications. Further research work is needed to explore its application beyond the equi-spaced residual points.
\section{Impact Statement}

The presented research makes a significant impact on the research area of physics-based machine learning techniques by introducing a novel algorithm for adaptive sampling of residual points. By addressing the challenges associated with existing adaptive sampling schemes, the proposed algorithm enhances prediction accuracy, computational efficiency, and training robustness. Through verification studies, the algorithm demonstrates reasonable agreement with benchmark model results, showcasing its effectiveness in improving simulation accuracy. Comparative analysis with other techniques underscores the superiority of the proposed approach, particularly with higher refinement factors.  

\bibliography{example_paper}
\bibliographystyle{synsml2023}

\section*{References}

\medskip

\small
[1] Karniadakis, G. E., Kevrekidis, I. G., Lu, L., Perdikaris,
P., Wang, S., and Yang, L. Physics-informed machine
learning. Nature Reviews Physics, 3(6):422–440, 2021.

[2]	Psaros, A. F., Kawaguchi, K., and Karniadakis, G. E. Metalearning
pinn loss functions. Journal of Computational
Physics, 458:111121, 2022.

[3] Yu, J., Lu, L., Meng, X., and Karniadakis, G. E. Gradientenhanced
physics-informed neural networks for forward and inverse pde problems. Computer Methods in Applied Mechanics and Engineering, 393:114823, 2022.

[4]	Wu, C., Zhu, M., Tan, Q., Kartha, Y., and Lu, L. A comprehensive
study of non-adaptive and residual-based adaptive
sampling for physics-informed neural networks. Computer
Methods in Applied Mechanics and Engineering,
403:115671, 2023.

[5] Lu, L., Meng, X., Mao, Z., and Karniadakis, G. E. Deepxde:
A deep learning library for solving differential equations.
SIAM review, 63(1):208–228, 2021.

[6]	Nabian, M. A., Gladstone, R. J., and Meidani, H. Efficient
training of physics-informed neural networks via importance
sampling. Computer-Aided Civil and Infrastructure
Engineering, 36(8):962–977, 2021.

[7]	Subramanian, S., Kirby, R. M., Mahoney, M. W., and
Gholami, A. Adaptive self-supervision algorithms
for physics-informed neural networks. arXiv preprint
arXiv:2207.04084, 2022.

[8] Daw, A., Bu, J., Wang, S., Perdikaris, P., and Karpatne,A. Rethinking the importance of sampling in physics-informed neural networks. arXiv preprint
arXiv:2207.02338, 2022.

[9]	Tang, K.,Wan, X., and Yang, C. Das-pinns: A deep adaptive
sampling method for solving high-dimensional partial
differential equations. Journal of Computational Physics,
476:111868, 2023.

[10]	Peng, W., Zhou, W., Zhang, X., Yao, W., and Liu, Z.
Rang: A residual-based adaptive node generation method
for physics-informed neural networks. arXiv preprint
arXiv:2205.01051, 2022.

[11] Gao, Z., Yan, L., and Zhou, T. Failure-informed adaptive
sampling for pinns. arXiv preprint arXiv:2210.00279,
2022.

[12] Zeng, S., Zhang, Z., and Zou, Q. Adaptive deep neural
networks methods for high-dimensional partial differential
equations. Journal of Computational Physics, 463:
111232, 2022.

[13]	Nilabh, S. and Grandia, F. Dynamic weights enabled
physics-informed neural network for simulating the mobility
of engineered nano-particles in a contaminated
aquifer. arXiv preprint arXiv:2211.03525, 2022.

%
%

\end{document}